# Online programming system for robotic fillet welding in Industry 4.0

*Ignacio Diaz-Cano*
Department of Automatic, Electronic, Computer Architecture and Communication Networks Engineering, University of Cádiz, Cádiz, Spain

*Fernando M. Quintana*
Department of Computer Science and Engineering, School of Engineering, University of Cádiz, Cádiz, Spain

*Miguel Lopez-Fuster and Francisco-Javier Badesa*
Department of Automatic, Electronic, Computer Architecture and Communication Networks Engineering, University of Cádiz, Cádiz, Spain

*Pedro L. Galindo*
Department of Computer Science and Engineering, School of Engineering, University of Cádiz, Cádiz, Spain, and

*Arturo Morgado-Estevez*
Department of Automatic, Electronic, Computer Architecture and Communication Networks Engineering, University of Cádiz, Cádiz, Spain

**Abstract**

**Purpose** – Fillet welding is one of the most widespread types of welding in the industry, which is still carried out manually or automated by contact. This paper aims to describe an online programming system for noncontact fillet welding robots with "U"- and "L"-shaped structures, which responds to the needs of the Fourth Industrial Revolution.

**Design/methodology/approach** – In this paper, the authors propose an online robot programming methodology that eliminates unnecessary steps traditionally performed in robotic welding, so that the operator only performs three steps to complete the welding task. First, choose the piece to weld. Then, enter the welding parameters. Finally, it sends the automatically generated program to the robot.

**Findings** – The system finally managed to perform the fillet welding task with the proposed method in a more efficient preparation time than the compared methods. For this, a reduced number of components was used compared to other systems: a structured light 3 D camera, two computers and a concentrator, in addition to the six-axis industrial robotic arm. The operating complexity of the system has been reduced as much as possible.

**Practical implications** – To the best of the authors' knowledge, there is no scientific or commercial evidence of an online robot programming system capable of performing a fillet welding process, simplifying the process so that it is completely transparent for the operator and framed in the Industry 4.0 paradigm. Its commercial potential lies mainly in its simple and low-cost implementation in a flexible system capable of adapting to any industrial fillet welding job and to any support that can accommodate it.

**Originality/value** – In this study, a robotic robust system is achieved, aligned to Industry 4.0, with a friendly, intuitive and simple interface for an operator who does not need to have knowledge of industrial robotics, allowing him to perform a fillet welding saving time and increasing productivity.

**Paper type** Research paper

## 1. Introduction

Industrial robotics did not appear until the third Industrial Revolution. They represented a new paradigm around manufacturing, focusing it on a new way of increasing productivity in terms of lower costs and faster work speed. The inclusion of robots in the industry was taken as a new opportunity to automate all serial production, which had already appeared in the second Industrial Revolution. Also contributing to this is the parallel emergence of new materials, more flexible and powerful software and more reliable electronic devices (Salisbury, 1980). In this third revolution, the first industrial robots were not intuitive at all, so their programming was extensive and complex and had to be carried out by a well-trained operator (E.I, 2013; Tzvetkova, 2014).

Nowadays, companies are still looking for improvements in their manufacturing process. With the appearance of the concept of Industry 4.0 throughout 2011 (Lu, 2017), all

The current issue and full text archive of this journal is available on Emerald Insight at: https://www.emerald.com/insight/0143-991X.htm

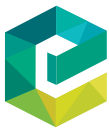

Industrial Robot: the international journal of robotics research and application
© Emerald Publishing Limited [ISSN 0143-991X]
[DOI 10.1108/IR-07-2021-0137]

This work was supported by the Spanish Plan estatal de Investigación Científica y Técnica y de Innovación 2017-2020 under contract (AUROVI) EQC2018-005190-P. Fernando M. Quintana would like to acknowledge the Spanish Ministerio de Ciencia, Innovación y Universidades for the support through FPU grant (FPU18/04321).

Received 6 July 2021
Revised 24 August 2021
21 September 2021
Accepted 27 September 2021

industrial sectors have set out to achieve the objectives proposed by this paradigm, called the Fourth Industrial Revolution. To achieve the objectives proposed in the Fourth Industrial Revolution, up to nine key enabling (Laganière, 2011; Lu, 2017) technologies have been established. This work takes advantage of several of those key enabling technologies to implement an advanced approach for industrial welding using autonomous robots and taking shipbuilding as a case study.

Welding robots of the third industrial revolution, in general, were very large and heavy structures, inflexible and slow to program, which required the operator to spend a lot of time occupying the production area, with the consequent added cost (Kim, 2008). However, as we move into the 21st century, robots tend to be programmed off-line to better optimize production times achieving higher precision, an essential feature in shipbuilding welding (Ang *et al.*, 1999). Alternatively, they can be programmed online and minimize the track time, that is, the required product assembly duration that is needed to match the demand. Nowadays, computer vision is used to achieve, not only an efficient track time but also a better precision in the work to be done. The authors show different applications and vision technologies in Bogue (2018) based on a traditional sequence of steps and on the laser as a vision device.

In the last year, various approaches to robotic vision welding have been reviewed, updated and proposed. Thus, in Lei *et al.* (2020), the authors point out as a challenge the positioning of the welding start point for efficient online robot work, proposing a complex achievement of detection algorithms or specific programming. The authors in Shen *et al.* (2020) propose a data model for automated welding, relying on CAD/computer-aided manufacturing/computer-aided process planning technology, achieving a robust system but with an excess of devices and steps in their method that could be solved by taking another approach. The authors in Chen and Hu (2021) show a specific double-arm welding robot focused on stud welding, where the system relies on a computer-aided design (CAD) model to develop the welding trajectories. On shipbuilding, the authors analyze in Zych (2021) current programming methods, where they dedicate a space to the use of vision in ship welding. As a future analysis, they propose a study of the application of 3D sensors in the capture of the scene in real time, which is addressed in the present work.

Therefore, current robot programming can be classified into two main categories: online and offline. In online programming, in the most basic way the operator controls the robot from the teach pendant (TP), moving the robot to each point. These points are stored in a program within the robot itself and are run when needed (Pan *et al.*, 2010). There are also other customized solutions called operator-assisted online programming, where the robot is controlled through a human–machine interface (HMI). On the other hand, offline programming is based on semi-automated platforms, which need to incorporate precise CAD designs of the scene to be simulated. In this way, it is possible to emulate manufacturing tasks outside the production area. This tends to avoid long and costly tasks within the line of work. The CAD designs used should reflect as accurately as possible the specific stage to be modeled, which often includes complex designs or robot calibrations. However, sometimes, the CAD designs are used in online programming robots systems (Bedaka *et al.*, 2019). In this work, we consider the step in which the CAD of the piece is compared with the real scene unnecessary, and therefore, it is eliminated.

To the best of our knowledge, there is no scientific evidence of an online programming system for robots capable of carrying out a fillet welding process, such as the one presented in this study. Thus, its main contributions are as follows:

- effective integration of various technologies to obtain a robotic fillet welding system with reliable and secure communications, efficiently distributing automated tasks between a computer, where the user operates, and a computer, which manages the movements of the robot, reducing the process time compared to previous methods;
- an appropriate workflow adapted to the proposed system, eliminating unnecessary steps in the welding process. An example is the comparison with a three dimensional model of the structure, as the scene is captured live at all times; and
- the creation of an intuitive, specific and simple HMI, greatly reducing the time to start the welding process.

Thus, the article is organized as follows. Section 2 shows the proposed system overview. Section 3 presents the methodology followed in this study. Section 4 collects the experimentation and results. Finally, in Section 5, the conclusions and future work are presented.

## 2. System overview

An environment has been designed where two computers communicated with each other through an access point, together with a robotic arm, share the performance of all the tasks involved in the welding process. The system is made up of three different parts that will be described below. A diagram of the architecture of the proposed system is shown in Figure 1.

### 2.1 Robot side

The *robot side* is the one near the scene to capture. In it, a high-resolution structured light camera takes care of the capture, connected via USB 3.0 to the robot handler (a camera connected via Ethernet might also be considered). The camera is anchored to the robot, on its right side, jointly, so that when the robot moves, the camera moves with it, as shown in Figure 2. The robot handler is responsible for calibrating and aligning the camera with respect to the robot. The welding wire feeding machine is anchored to the left side of the robot.

The *robot handler* is a computer responsible for the communication of the operator HMI (*operator side*) with the robot, through the (*communications*) that have been established. In turn, the robot handler communicates with the robot and transfers all the requests that come to it from the HMI through a socket and Robot Operating System (ROS) (ROS.org, 2021). Communication with the camera, obtaining the position of the robot, the state in which it is at all times, is achieved through the compatibility that it also has with ROS. When requested, the information is transferred to the operator according to the established protocol.

### 2.2 Communications

Communication between the robot handler and the robot is required to be able to control its position and performance.

**Figure 1** Diagram of the proposed system architecture

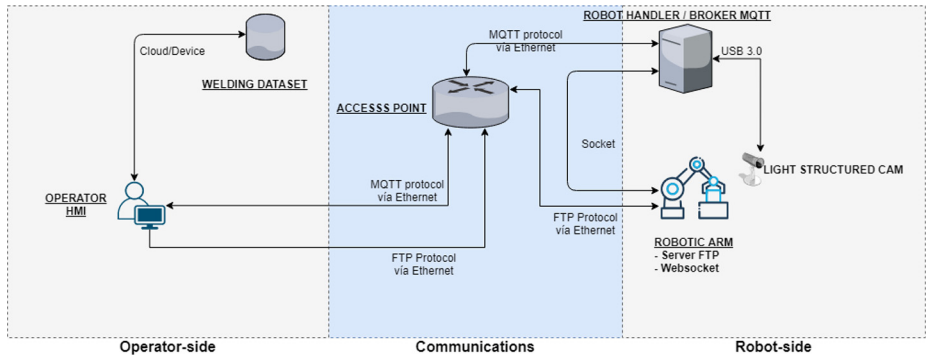

**Figure 2** Robot Fanuc LR Mate 200 id/7L with Zivid One+ M structured light cam anchored

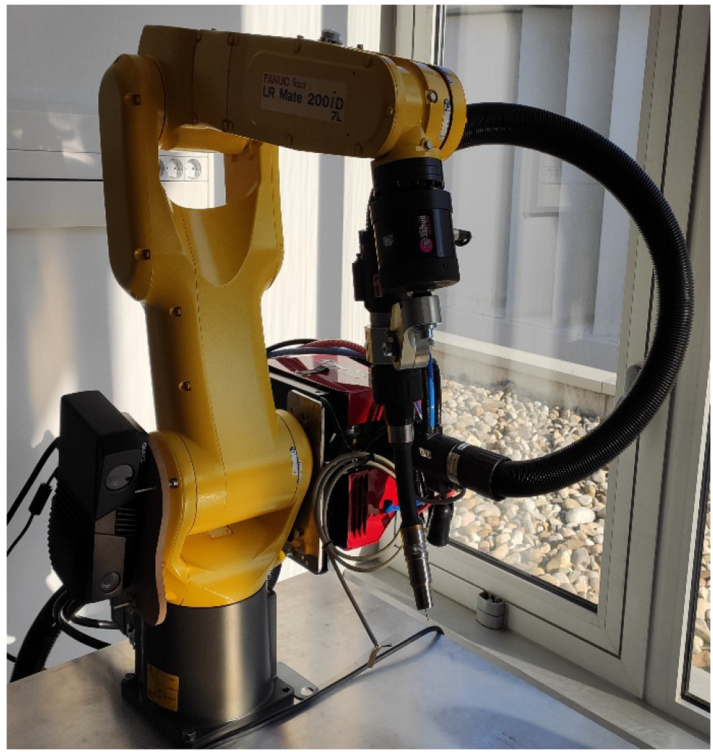

Likewise, there must be direct communication with the operator, as he will generate the welding program and send it to the robot. Finally, a general communication between the customer and the robot is needed to send the different orders that make up the steps of the welding process.

To respond to communication needs, different network communication protocols have been chosen. In the first place, for direct communication between the robot and the robot handler, in order for it to control its pose, is done through a socket that the robot itself has implemented. Finally, to achieve fluid communication during all the steps that make up the welding process and so that the customer knows the situation of the robot at all times, it has been decided to implement the message queuing telemetry transport (MQTT) protocol.

As the authors explain in Hillar (2017), MQTT is a lightweight publish–subscribe network protocol transporting messages between different devices. MQTT has two entities within the network: broker and one or more clients. The broker is a software that acts as a server and the clients can be any device that connects to the network. Information is shared through messages across topics. When an element of the network (publisher) has an element to share on the network, it is sent to the broker, and this is sent to all clients that are subscribed to the topic for which the publisher has sent the message. Thus, the main advantages of the MQTT broker are as follows:
- Eliminate vulnerable and insecure client connections.
- You can easily scale from a single device to thousands.

- Manage and track all client connection states, including credentials and security certificates.
- Reduce network tension without compromising security (cellular or satellite network).

This protocol is often widely used for communication on internet of things (IoT) platforms, as stated in MQTT (Saxena *et al.*, 2019). The advantages it offers, together with its operation, make it ideal for the architecture presented in this study, as it meets what is required to achieve the objectives of a communication in a project that meets the Industry 4.0 standard.

However, both this computer and the operator (HMI) act as clients and servers within the MQTT network, as both sides are subscribed and publish on the same topic where messages are shared according to the task to be performed. Once an element of the system publishes a message associated with a topic, all the elements that are subscribed to that topic will receive the message and perform the function that is programmed in each case.

It should be noted that the proposed network is not part of the factory's corporate network but is isolated and external to it without having an internet connection.

However, all file transfer protocol and MQTT communications are carried out using secure sockets layer (SSL) encryption. In the listing, you can see an example of connection to the MQTT broker using SSL. In Listing 1, you can see an example of the code sequence necessary for communications in MQTT to be carried out securely using SSL.

The entire process is done by sending and receiving messages between the robot handler and the operator (HMI) by MQTT. Table 1 shows the list of messages that are exchanged between the two computers that control the system: Robot Handler and Operator (HMI), during the welding process. In each row of the table you can read the internal operating command, transmitter and receiver of the message, as well as a description of it.

*Listing 1. System messages under MQTT protocol*

```
QSslConfiguration sslConfig = QSslConfiguration::defaultConfiguration();
//Add custom SSL options here (for example extra certificates)
QMQTT::Client *client = new QMQTT::Client("IP\_broker", 8883, sslConfig);
client->setClientId("clientId");
client->setUsername("user");
client->setPassword("password");
//Optionally, set ssl errors you want to ignore. This may weaken security. QSslSocket::ignoreSslErrors(const QList<QSsError>\&)
client->ignoreSslErrors(<list of expected ssl errors>);
client->connectToHost();
```

### 2.3 Operator side

On the *operator side* is the operator, who communicates with the system through software containing an intuitive HMI. For its design, it has had the help of an experienced naval welder, who has explained to us the steps that a human would have to take to carry out the complete welding process.

Based on the instructions of the welding expert, a system has been created where the operator only has to interact three times to start the welding process, which is shown in Figure 3. First step, the operator selects a type of structure to be welded. This choice will send the robot handler a message for the camera to start capturing the piece. Once this capture is completed, it will be sent through the communications channel and will appear on the left side of the screen (real image), in the form of a 3D point cloud, along with their dimensions. A diagram of the structure to be welded (scheme) will appear on the right, together with the maximum measurement of the horizontal and vertical part of the structure. This choice will send the robot handler a message for the camera to start capturing the piece.

**Table 1** System messages under MQTT protocol

| Command | Transmitter | Receiver | Description |
|---|---|---|---|
| "InterfaceReady" | Operator(HMI) | Robot handler | The operator application informs the robot handler that it is in the welding panel and ready to start work |
| "HandlerRobotReady" | Robot handler | Operator(HMI) | The robot handler performs the necessary checks to be able to start the work: connection with the robot and the camera, ok |
| "Capture" | Operator(HMI) | Robot handler | The operator requests the robot handler to send an order to the camera to capture the scene |
| "AnswerCapture" | Robot handler | Operator(HMI) | Once the camera has captured the scene, the robot handler sends a JSON structure with the data necessary to perform the welding |
| "FTP_OK" | Operator(HMI) | Operator(HMI) | The operator sends this message if the robot correctly loads the welding program that the operator has sent via FTP. The HMI receives it and informs the operator of it |
| "FTP_NO_OK" | Operator(HMI) | Operator(HMI) | The robot sends an incorrect load signal of the program, and the operator informs the operator of this fact to restart the process |
| "Welding" | Operator(HMI) | Robot handler | Once the welding program has been sent to the robot, the HMI asks the robot handler to order the robot to start the sent weld |
| "EndWelding" | Robot handler | Operator(HMI) | The robot manipulator informs the operator, via a message that appears on the HMI, that the welding is complete |
| "Pickup" | Operator(HMI) | Robot handler | The operator asks the robot handler to command the robot to return to the home position to start another weld |
| "Pickuped" | Robot handler | Operator(HMI) | The robot handler informs the operator that the robot has already reached the start position so that the operator can start another weld again |

**Figure 3** User-controlled HMI application showing the consecutive steps involved in welding process (1) selecting the structure to be welded; (2) choosing the appropriate welding parameters; and (3) sending the order for welding. A communication log is also shown for sending messages to the user to assess its interaction with the system

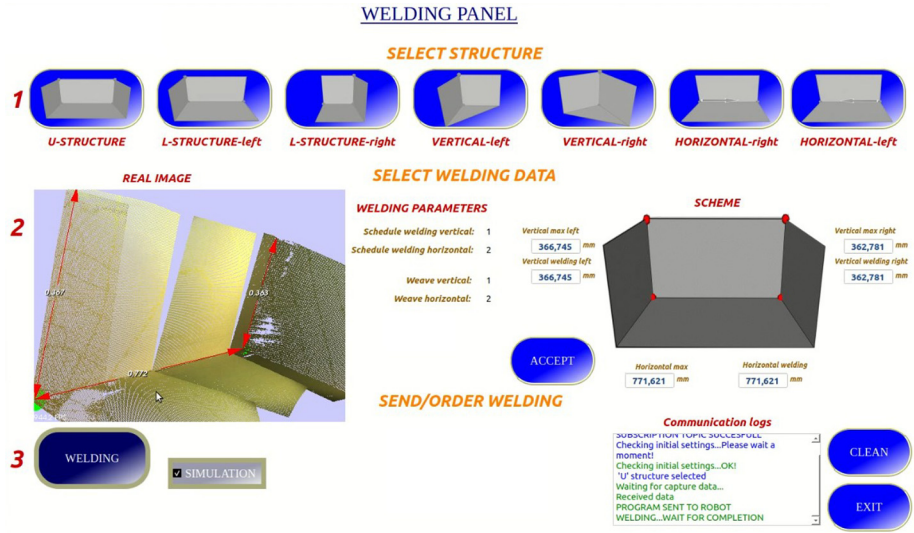

Next, the second step the operator must select the length to be welded below the maximum of the system measured. Finally, you must choose the welding and weave sine schemes that will be applied to the job. Both are stored in the robot, this interface only makes a call to them.

Finally, finally in the third step, the operator must choose if he wants to carry out a simulation of the welding, with which the robot will only carry out the necessary trajectories to achieve the final work but without lighting the torch and therefore without carrying out the welding, or it will carry out the complete welding. Initially, the program will be sent to the robot, already generated, depending on the structure chosen at the beginning, and then the operator will give the order to the robot handler so that it tells the robot to start welding. At the end of it, the robot handler will notify the operator HMI that the job has been concluded and the form will be restarted to accept a new interaction.

Once this capture is completed, it will be sent through the communications channel and will appear on the left side of the screen (real image), in the form of a 3 D point cloud, along with their dimensions. A diagram of the structure to be welded (scheme) will appear on the right, together with the maximum measurement of the horizontal and vertical part of the structure. In this interaction, the operator will be able to see the scanned part from a different position and must select the length to be welded below the maximum that the system has measured. Finally, you must choose the welding and weave sine scheme to be applied to the job; both are stored in the robot, and this interface only makes one call to them.

Second, and after having sent a message to the robot handler with the choice of the structure, the camera will start capturing the piece. Once this capture is completed, it will be sent through the communications channel and will appear on the left side of the screen (real image), in the form of a 3 D point cloud, along with their dimensions. A diagram of the structure to be welded (scheme) will appear on the right side, together with the maximum length of the horizontal and vertical part of the structure. In this interaction, the operator will be able to see the scanned part from a different position and must select the length to be welded over the maximum that the system has measured. Finally, you must choose the welding and pendulum scheme to be applied to the job; both parameters are stored in the robot, and this interface only makes one call to them.

Third, the operator must choose if he wants to carry out a simulation of the welding, with which the robot will only carry out the necessary trajectories to achieve the final work but without lighting the torch and therefore without carrying out the welding, or it will carry out the complete welding. Initially, the program will be sent to the robot, already generated, depending on the structure chosen at the beginning, and then the operator will give the order to the robot handler so that it tells the robot to start welding. At the end of it, the robot handler will notify the operator HMI that the job has been concluded and the form will be restarted to accept a new interaction.

## 3. Methodology

Industrial robots are not very intuitive to use and are of considerable complexity, especially if they are to be used in precise tasks with complex geometries such as ship welding (Stanić *et al.*, 2018). That is why an interface is required that acts as a simple and intuitive HMI, as transparent as possible, so that an operator, expert in welding, who does not know how the programming of a robot works, is able to manage and communicate with it.

The proposed method is based on a classic computer-assisted online robot programming system for welding tasks, as proposed by Guhl *et al.* (2019), but adapting and optimizing the key steps of the method. In this way, the aim is to produce weld seams in the in the shortest possible time and with the greatest possible efficiency. In the proposed case study, the shipbuilding, is governed by a precision error determined by ISO 5817 (ISO, 2014) taken into account in this study.

Figure 5 shows the sequence diagram that has been proposed and which optimizes the conventional programming method on which it is based. You can also see the place where each of the system messages in Table 1 is executed. This sequence diagram shows a valid route planning transaction with no errors. Each of the tasks performed by the proposed system is detailed below, locating each of the key steps of the original method.

### 3.1 Calibration tool

The first step to take when using any tool, only once, before you start programming the robot or when the tool suffers any alteration in its position because of improper movement, shock or displacement in general. The tool in the proposed case study is a welding torch. All robot manufacturing companies use very similar methods. Their methods are based in the same idea:

- Calculate the position of tool center point (TCP).
- Calculate the rotation especially for complex tools or those that have an angle on it.

In the case of Fanuc (robot used in the study), they use the method known as the three-point method for a simple tool and the six-point method for a complex tool as this is our case. Therefore, we need to calculate not only its position but also its orientation. Three points are calculated in space, as in the three-point method, and additionally the following points must be taken:

1. Orient origin point: This point corresponds to the origin of the orientation.
2. X direction point: With this point, we define the direction of the X-axis moving the robot in world coordinates.
3. Z/Y direction point: The last point is the one that corresponds to the Z or Y direction. Once we capture these six points, the robot calculates the Denavit–hartenberg parameters of the position and orientation of the tool.

### 3.2 Human–machine interface interaction/calculations

The key step in the interaction of the operator with the system is carried out in the simplest way possible in three steps. These steps are distributed throughout the process that were commented on in Section 3. It is a transversal step that will comprise an important part of the operator's interaction with the system.

Next, in the *calculations*, the robot handler will instruct the structured light camera to take a photo of the scene, obtaining a three-dimensional point cloud. Next, the random sample consensus (RANSAC) algorithm (Derpanis, 2010) is used to equations of the planes present in the scene. By intersecting these planes, the key points of the scanned structure are calculated.

In the case of fillet welding, the key points will be those that correspond to the corners and the consequent intersection of three planes. Once these points have been found, and knowing the plans that make up the piece, it can be extrapolated to the remaining points knowing the distance at which they are. To do this, we calculate the normalized vector corresponding to the intersection of two planes, taking as origin the corner corresponding to the intersection of the three planes and multiplying the vector by the distance at which the point is located. The Figure 4(a) shows a real L-shaped structure, while Figure 4(b) shows a 3D model of the same structure, where the three planes detected are observed in different color: red, green and blue.

Finally, the remaining points are obtained, a procedure suggested by the authors in Yang and Förstner (2010). The inclination of the tool at each point is also calculated, according to the ISO 6947 (ISO, 2019) standard, applied by the authors in Li *et al.* (2016). Subsequently, all the information will be stored on the robot handler itself and will be sent through the MQTT protocol with the message "AnswerCapture" in a javaScript object notation structure to the operator HMI.

The robot handler controls the pose of the robot and the camera through ROS (Zivid, 2020; ROS.org, 2021); the rest of the process is implemented in C++ language, additionally using the point cloud library (PCL) library (Rusu and Cousins, 2011).

### 3.3 Create/send/execute program

Once the operator has chosen the welding parameters and the points where the welding paths are to be planned, he will issue the order to weld. At that time, the operator application will generate the welding program dynamically, depending on the choice of the operator.

Next, the operator first sends the welding program to the robot, and if it arrived correctly, the interface will be enabled so that the operator can send the welding order to the robot handler,

**Figure 4** An example of an L-structure capture using the 3 D camera: (a) real image L-structure; (b) three 3 D cloud points showing in three different colors the planes identified by RANSAC algorithm and the L lines and the vertex found by intersections of the fitted planes

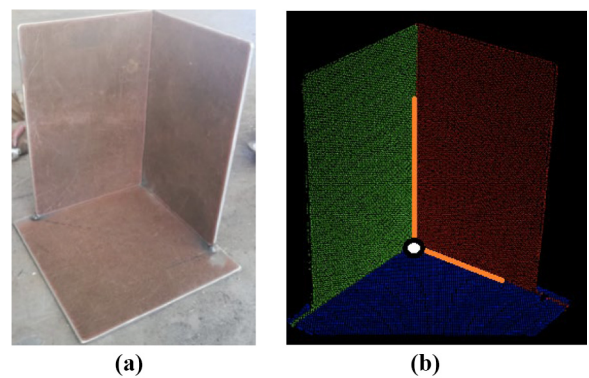

**Notes:** (a) Real image "L"-structure; (b) 3D point cloud – "L"-structure

**Figure 5** Diagram of the key steps of the proposed programming method

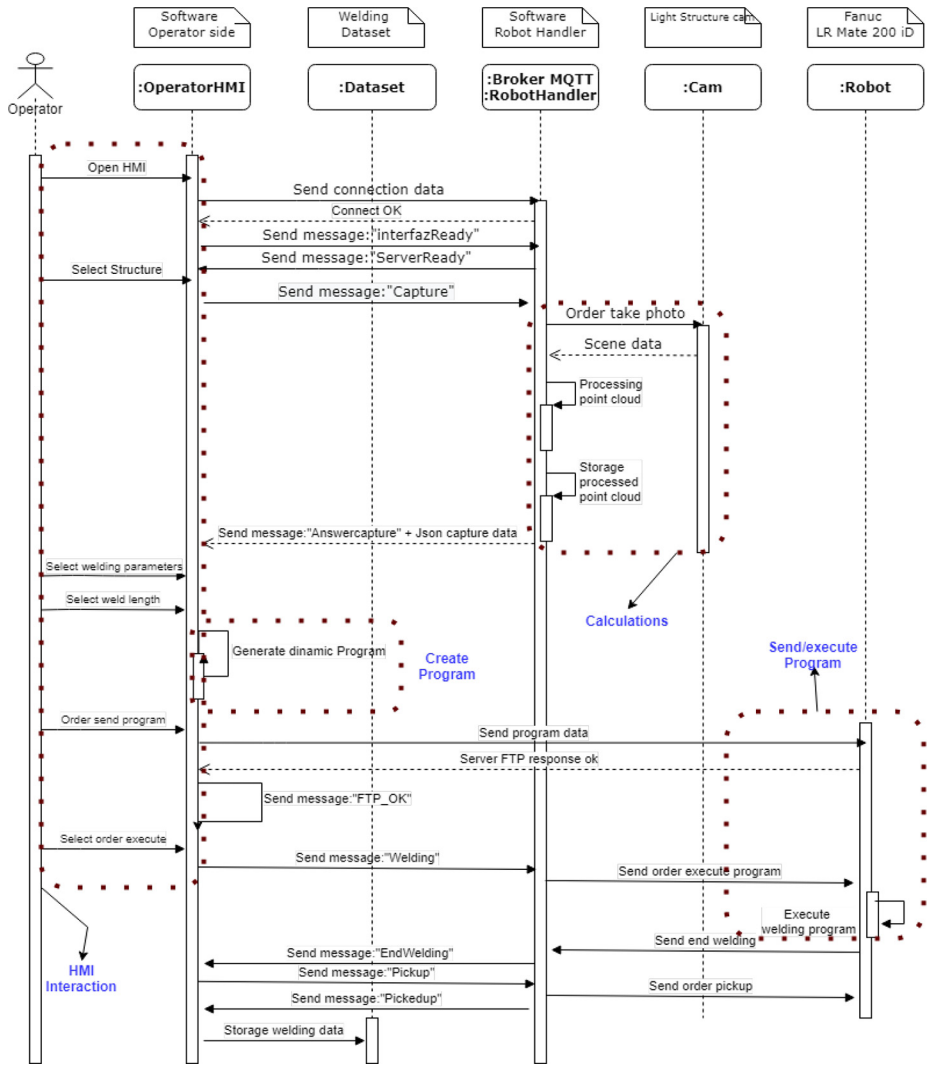

**Table 2** Stored data of each weld performed by the robot

| Data stored | Description |
| --- | --- |
| Timestamp | Date and time the work is done |
| Vertical welded distance | Distance, in centimeters, welded on the vertical |
| Horizontal welded distance | Distance, in centimeters, welded on the horizontal |
| Vertical maximum size | Maximum distance, in centimeters, from the vertical |
| Horizontal maximum size | Maximum distance, in centimeters, from the vertical |
| Process time | Total process time, from when the operator begins to interact with the system, until welding begins |
| Capture time | Time spent by the system in capturing the structure |
| Structure type | The type of structure that has been worked on is stored |
| Welding scheme | The number of welding scheme used in each job is stored |
| Weave sine scheme | The number of weave sine scheme used in each job is stored |

and the latter transfers it to the robot, executing the welding program.

### 3.4 Storage welding data

Finally, automatically and completely transparent to the operator, the operator application stores locally a record of the welding execution. The information is stored in ".csv" format so that any external application could connect with this source of information and transmit it to an IoT platform or carry out any other computing in the cloud (Figure 5).

To preserve the local network of the proposed system from cyberattacks, no type of internet connection is made from the operator application, which could be used to carry out an intrusion into it, but the data are left prepared in an external storage for any other application needing it. Thus, the stored data could be processed later if you wanted to store it on an IoT platform. The structure of the stored data are shown in the Table 2.

## 4. Experimental results

To validate the suitability of the proposed methodology and system architecture and to evaluate and compare online and offline robot programming approaches, the experimental environment and the results obtained are described and discussed below.

### 4.1 Experimental environment

The experimental environment was designed to implement the proposed methodology. Table 3 shows the components of the system architecture, which has been tested to plan welding trajectories in one of the most numerous repeatable structures within shipbuilding, U-shaped, also called open blocks (Lee, 2014).

To simulate the necessary trajectories in offline programming methods, a scenario has been generated according to the occasion and based on it. Thus, in Figure 6(a), you can see the scenario recreated for Roboguide (Li *et al.*, 2007), and in Figure 6(b), you can see the scenario generated for RoboDK (Ribeiro *et al.*, 2019).

In the experiment, the welding program of the mentioned part has been created in Figure 7. Thus, in Figure 7(a), you can see the real view of the piece, whereas in Figure 7(b), you can see the drawings, with the dimensions of the structure and the paths that will be generated and numbered in the order in which they have to be programmed. First, the horizontal edge will be welded and then the vertical one.

*Listing 2. Welding program code in a Fanuc robot for a U-shaped structure*

```
 1: J P[1] 10% FINE;
 2: L P[2] 100 mm/sec FINE;
 3: L P[3] WELD_SPEED CNT100;
 4: L P[4] 100 mm/sec FINE;
 5: L P[3] 100 mm/sec FINE;
 6: L P[5] WELD_SPEED CNT100
 7: J P[6] 100 mm/sec FINE;
 8: L P[7] 100 mm/sec FINE;
 9: L P[8] WELD_SPEED CNT100;
10: L P[9] 100 mm/sec FINE;
11: L P[8] WELD_SPEED CNT100;
12: J P[10] 100 mm/sec FINE;
```

A fair comparison should be made between the four methods involved in the investigation. For this, the same welding program will be generated. Thus, in the program in Listing 2, the lines of code that should ignite the welding torch, and swing the robot (*weld_start* and *weave_sine* respectively), have been omitted. This decision has been made because its timing is constant and would not affect the timing of the experiment.

It has been seen that four paths are needed to weld this structure. Routes 1, 2, 3 and 4 according to Figure 7(b) are identified by instructions 3, 6, 9 and 11, respectively. The rest of the instructions of the program are points of approach and withdrawal of the robot necessary to complete the task.

**Table 3** Items that make up the system architecture

| Item | Description | Role |
|---|---|---|
| PC | 8th generation processor, 8 Gb RAM, 1 Tb SSD, Ubuntu 18.04 | Operator HMI |
| PC | 8th generation processor, 16 Gb RAM, 1 Tb SSD, Ubuntu 18.04 | Robot Handler |
| Switch | RJ-45 Ethernet switch: Gigabit Ethernet (10/100/1000), 8 ports. Full duplex | Access point |
| Cam | Camera Zivid One+ M. 0,1mm precision until 1.20 m. Figure 2 | light structure cam |
| Robot | Fanuc LR Mate 200 iD/7L. Figure 2 | Robotic arm |
| MQTT | Eclipse Mosquitto. An open source MQTT broker. See Eclipse (2021) | Broker MQTT |

**Figure 6** Offline programming scenarios for U-structure

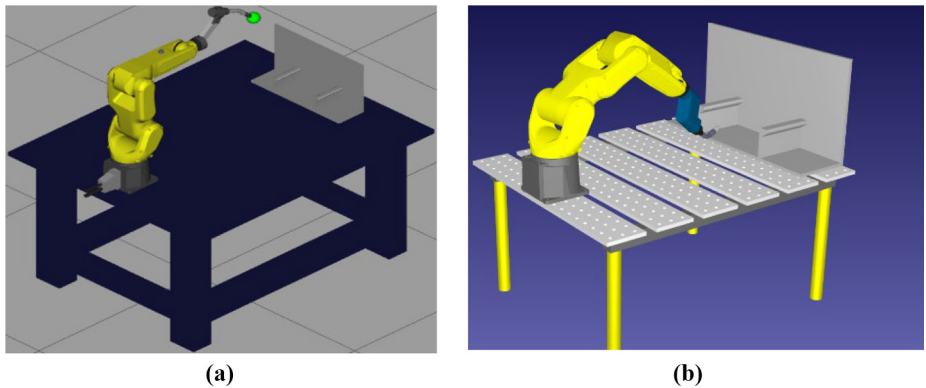

(a)      (b)

**Notes:** (a) Stage U-structure created by Roboguide; (b) stage U-structure created by RoboDK

**Figure 7** "U"-structure sample used for described welding experiment

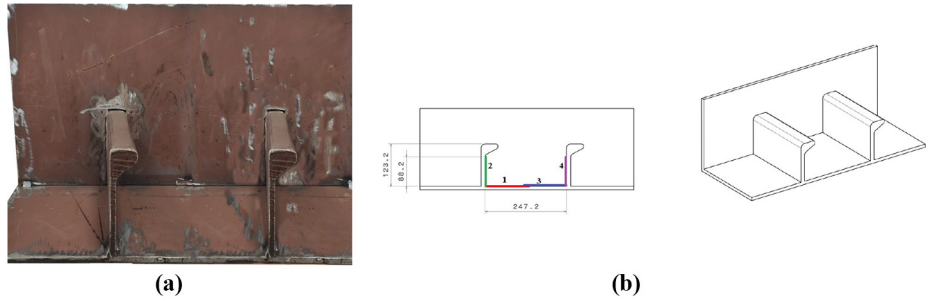

(a)                                           (b)

**Notes:** (a) "U"-structure picture taken in a real environment; (b) "U"-structure drawing

### 4.2 Results and discussion

In the experiment, the necessary trajectories have been generated before welding a U-shaped structure, numerous in shipbuilding. The experiment shows the time difference between the four programming options that have been proposed in this study (two of them online and two offline). For the study of repeatability, each of the simulations has been repeated five times for each programming option.

During the experiment, the execution times of each of the key steps that the authors propose in Guhl *et al.* (2019) for online programming methods. In the same way, the same operation has been carried out for the key steps that the authors propose in Larkin *et al.* (2011) for offline programming methods.

Due to all the executions were close to the mean value and there were no outline values, the mean of the times obtained as a result in each of the five execution repetitions of each method is taken as a measure. These data can be seen, numerically in Table 4 and graphically in Figure 8. Both results are shown in seconds.

The rows in the table show each of the programming methods involved. On the other hand, the columns represent each of the key steps of each method and the total time. Thus, the table is separated, so that the upper part shows the results of the offline programming methods, whereas the lower part of the table shows the results of the online programming methods.

The data are shown graphically in two figures. In each of them, the X axis represents the key steps of each programming method, while the Y axis shows the time invested in seconds.

If we look at the offline methods, it is observed that there is hardly any difference between one type of software and another, the generic framework (RoboDK) being somewhat faster. Therefore, in case of opting for offline programming, the software package used would not matter. For online programming methods, the authors of Guhl *et al.* (2019) claim that they are suitable for high volumes of repetitive work. However, the occupation time in the production area must be minimal. Online programming by the simplest method, the TP, has been carried out by an experienced welder with a high qualification in handling robots. However, it is observed how the time it takes to generate a program is too long, in comparison with the proposed method and with offline methods.

A comparison has also been made with two commercial solutions, taking the data from their Web pages, as there are no papers that show the operating data. In the first one, Inrotech (2021) has proven that its laser-based vision system has a total operating time that amounts to almost double that of the proposed method. In the second commercial (Pemamek, 2021), a system can be seen that is capable of performing more functions than the proposed method but that requires more in-depth training, as well as a more complex handling than the proposed method. In addition, neither of the two systems is capable of preserving the data of each job to later draw conclusions or apply data analysis techniques that allow us to improve the environment.

Regarding the implications of the study, one can speak of a theoretical implication, in the sense that a method and a series of constituent elements have been proposed that align any project to Industry 4.0. On the other hand, one can speak of a practical implication, in the sense that the system can be easily matured to be able to be commercialized with a low cost and a very efficient result.

**Table 4** Planning execution time by key step

| Programming Methods | 3d model Generation | Tag Generation | Trajectory Planning | Process Planning | Post Processing | Robot Program | Total Time |
|---|---|---|---|---|---|---|---|
| | | | Offline Programming | | | | |
| RoboDK | 131 | 28 | 23 | 13 | 22 | 24 | 241 |
| Roboguide | 138 | 29 | 22 | 14 | 23 | 21 | 247 |
| | | | Online programming | | | | |
| Teach pendant | TP/HMI Interaction | Calculations | Creat program | Send/Execute program | | Total time | |
| Proposed | 36 | 33 | 550 | 5 | | 624 | |
| method | 52 | 10 | 4 | 7 | | 73 | |

**Figure 8** Planning execution time by key stepNotes: a. Online methods; b. offline methods.

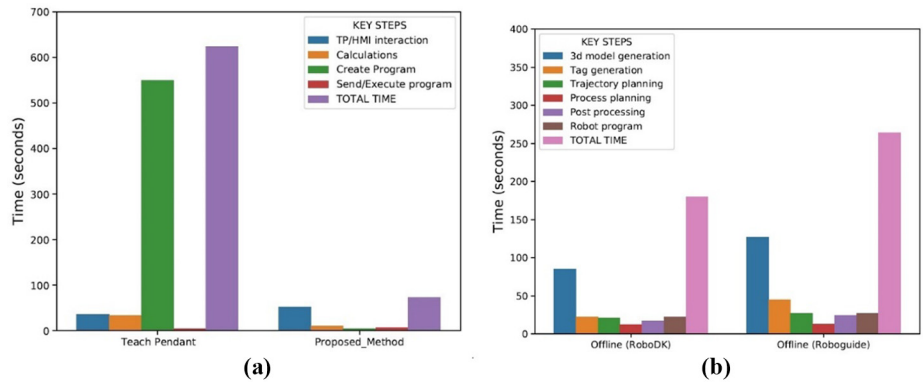

**Notes:** (a) Online methods; (b) offline methods

## 5. Conclusions

This paper presents an online programming system robotic for fillet with a hardware architecture capable of controlling the robot from an intuitive and easy-to-use interface that responds to some of the needs of the Fourth Industrial Revolution, incorporating several enabling technologies of this paradigm: *cybersecurity* (MQTT communications with SSL); *systems integration* (ROS systems have been integrated with C ++ programming and libraries such as virtual tool kit and PCL); *autonomous robots* (the robot has been endowed with the greatest autonomy with a control simple from the operator panel); *process simulation* (each welding task can be previously simulated); and *data generation* for a possible cloud computing platform or IoT (the data of each welding are saved in a data set to be processed in the future).

In addition to the successful incorporation of these enabling technologies of Industry 4.0, as reflected in Schiele *et al.* (2021), we have also had taken into account the ethical implications and the opinions of the operators to be able to complete a simple welding model in three steps and to contribute all the welding concepts to the proposed system, giving importance not only to the technological ones but also to the ethical and social implications.

It has been shown how the proposed method, based on operator-assisted online programming, requires less processing time than the rest of the methods currently used in robot programming, even from other commercial solutions created specifically for robotic fillet welding. However, it is found that the welding times of offline programming methods are acceptable, but it must be remembered that to carry them out, an experienced programmer is needed, in addition to the welder who supervises the welds carried out with the consequent extra expense of personnel.

In addition, the elimination of unnecessary steps has been successfully achieved, such as the comparison of the scene captured in each welding with the CAD design previously taken, as other systems use (Shen *et al.*, 2020). The remaining key steps are implemented under an abstraction layer for the operator, who only has to select three simple options to complete the welding task: structure, select welding parameters and send program automatically based on the first two options.

The position of the initial welding point, an important element to achieve (Lei *et al.*, 2020), is previously determined, according to the welding instructions implemented in the operator software at the expense of the indications of this.

The system has the limitation that it is not capable of performing any task if it does not find three planes, that is, if it is not capable of detecting a "U" structure or an open structure. If it is, this structure can address the welding of any of its parts, but it would not work in another situation. Nor is it capable of welding other structures other than the joints between planes, and its work area is limited to a horizontal 90 cm × 70 cm vertical, as the robotic arm does not advance during the welding task.

Additional work needs to be done to assess the potential benefits of incorporating other enabling technologies that make up Industry 4.0. Likewise, more case studies could be proposed for other productive sectors other than shipbuilding where fillet welding work is carried out.

However, the expansion of the system will be limited by the bracket on which it is installed. Thus, if it is intended in the future to expand the number of structures to be welded, in terms of dimension and shape, a preliminary study should be carried out on the convenience of this extension or the realization of a different system.

## Corresponding author

**Ignacio Diaz-Cano** can be contacted at: ignacio.diaz@uca.es